\documentclass{article}

\usepackage[utf8]{inputenc}
\usepackage{spconf}
\usepackage{mathtools}
\usepackage{tcolorbox}
\usepackage{amsmath}
\usepackage{amssymb}
\usepackage{amsfonts}
\usepackage{mathtools}
\usepackage{times}
\usepackage{tikz}
\usepackage{enumitem}
\usepackage{algorithm, algpseudocode}
\usepackage{multirow}
\usepackage{graphicx}
\usepackage{subcaption}
\usepackage{float}

\input{mysymbol.sty}
\input{mysymboltensor.sty}

\title{Tensor Low-rank Approximation of Finite-horizon Value Functions}

\name{\vspace{-.25cm}Sergio Rozada, and Antonio G. Marques \thanks{ Work supported by the Spanish NSF (AEI/10.13039/501100011033) grants PID2019-105032GB-I00, TED2021-130347B-I00, \&  PID2022-136887NB-I00.
The authors are with the Dept. of Signal Theory and Comms., King Juan Carlos University, Madrid, Spain. Contact author: antonio.garcia.marques@urjc.es.} 
}
\address{Dept. of Signal Theory and Communications, King Juan Carlos University, Madrid, Spain}

\begin{document}
\maketitle

\ninept

\begin{abstract}%
The goal of reinforcement learning is estimating a policy that maps states to actions and maximizes the cumulative reward of a Markov Decision Process (MDP). This is oftentimes achieved by estimating first the optimal (reward) value function (VF) associated with each state-action pair. When the MDP has an infinite horizon, the optimal VFs and policies are stationary under mild conditions. However, in finite-horizon MDPs, the VFs (hence, the policies) vary with time. This poses a challenge since the number of VFs to estimate grows not only with the size of the state-action space but also with the time horizon. This paper proposes a non-parametric low-rank stochastic algorithm to approximate the VFs of finite-horizon MDPs. First, we represent the (unknown) VFs as a multi-dimensional array, or tensor, where time is one of the dimensions. Then, we use rewards sampled from the MDP to estimate the optimal VFs. More precisely, we use the (truncated) PARAFAC decomposition to design an online low-rank algorithm that recovers the entries of the tensor of VFs. The size of the low-rank PARAFAC model grows additively with respect to each of its dimensions, rendering our approach efficient, as demonstrated via numerical experiments.
\end{abstract}

\vspace{-.1cm}

\section{Introduction} 
\label{S:introduction}
\vspace{-.1cm}
Dynamical systems are becoming increasingly intricate, reflecting the multifaceted challenges of our modern world. The  complexity of these systems calls for advanced solutions that can autonomously navigate, control, and optimize their behavior. Consequently, there is a need for new strategies able to not only cope with but also leverage this complexity. In this context, Reinforcement Learning (RL) has emerged as a promising tool to learn how to interact with the world (environment) via trial and error \cite{sutton2018reinforcement, bertsekas2019reinforcement}. RL aims to solve sequential optimization problems. Typically, the environment is represented as a set of states, and within each state, agents can take an action from a set of possible actions. The quality of each state-action pair is measured as a numerical signal, known as reward. The goal in RL can then be summarized as deciding which is the best action to take in each state so that the reward aggregated over time is maximized.

More formally, the problems described above are usually modeled as a Markov Decision Process (MDP) \cite{puterman2014markov}. Learning in the MDP context is about estimating a policy, which is a function that maps the states of the MDP to actions. One common approach is to estimate first the value function (VF) associated with each state-action pair, to then infer a policy by maximizing greedily with respect to the actions. This set of approaches receives the name of value-based methods \cite{sutton2018reinforcement}. When the MDP is infinite-horizon, one can show that, under mild conditions, policies (hence, VFs) are stationary (i.e., time-independent) \cite{bertsekas2012dynamic}. However, in many practical setups, there is a maximum stopping time, so that decisions can only be made during a finite horizon (FH). When the MDP is FH, policies and VFs are time-varying~\cite{puterman2014markov}. In other words, even if in two different time instants (say the initial one and the final one) the state is the same, the best action (response) to that state in the initial time instant is likely to be different from the best action in the final stage. The fact of the VF being different for each time instant entails that the degrees of freedom (DoF) of an FH VF is $T$ times larger (with $T$ denoting the horizon) than the DoF of its infinite-horizon counterpart. This can be challenging since the number of state-action pairs for most practical problems is substantially large, and, as a result, each VF itself is difficult to estimate. To alleviate this (curse of dimensionality) problem in infinite-horizon problems, VF approximation schemes have been proposed \cite{bertsekas1996neuro}. These typically include neural networks (NNs), and linear models \cite{mnih2015human, lagoudakis2003least}. However, VF approximation has not been thoroughly studied in FH problems. Motivated by recent advances in low-rank value-based RL in infinite-horizon problems \cite{rozada2023tensor}, \emph{this paper} aims to design a \emph{stochastic tensor low-rank estimation} scheme for FH MDPs that is both generic enough to estimate the optimal VFs and efficient in terms of parameters. Sec. 2 introduces the notation used in FH RL problems. Sec. 3 describes our proposed tensor low-rank approach for FH RL problems. Sec. 4 shows the empirical performance of our algorithm in two scenarios, and Sec. 5 provides concluding remarks.

\vspace{.5mm}

\noindent \textbf{Related work and contribution.} VF approximation has been extensively studied in infinite-horizon problems \cite{bertsekas1996neuro, sutton2018reinforcement, geist2013algorithmic}. In the context of FH problems, NN-based approaches have been proposed in  (model-based) optimal control to approximate the VFs \cite{guzey2016neural, hure2021deep, zhao2014neural}. In (model-free) RL setups, fixed-horizon approaches have been introduced to stabilize the convergence of value-function approximation algorithms. Fixed-horizon methods approximate infinite-horizon problems by FH ones, where linear and NN-based function-approximation have been considered \cite{de2020fixed, dann2015sample}. Low-rank optimization has been widely studied in tensor approximation problems \cite{sidiropoulos2017tensor, kolda2009tensor, gandy2011tensor}, but literature on low-rank RL is scarce. There are recent works on value-based infinite-horizon VF approximation, using low-rank matrix \cite{yang2019harnessing, shah2020sample, rozada2021low} and tensor models \cite{rozada2023tensor, tsai2021tensor}. Lastly, in the context of optimal control of ODEs,  \cite{oster2022approximating} proposes a least-squares tensor regression to estimate the VFs of an FH problem, where the coefficients of the linear approximation are modeled as a tensor. This paper introduces a \emph{low-rank} design for value-based methods in FH problems via \emph{tensor} completion. More precisely, i) we consider value-based \emph{model-free} FH problems; ii) we model the (time-dependent) VFs as tensors; iii) we leverage low-rank via the \emph{PARAFAC} decomposition; and iv) we estimate the FH VFs by solving an \emph{online stochastic} tensor-completion problem. 

\section{Fundamentals of RL and FH problems}
\label{S:fundamentals}
\vspace{-2mm}
In RL, an agent interacts sequentially with the environment, modeled as a closed-loop setup. The environment is defined by a set of states $\ccalS$, a set of actions $\ccalA$, and a reward function $R^a_s = \mathbb{E}\left[ r | s, a \right]$ that quantifies the expected \emph{instantaneous} value of a given state-action pair $\left( s, a\right)$. In FH problems, we consider a discrete time-space of finite duration $\ccalT=\{1,...,T\}$, where each element $t \in \ccalT$ is a time index, and $T$ is referred to as \emph{time horizon}. In time-step $t$, the agent observes the current state $s_t$, and selects an action $a_t$. The environment transitions into a new state $s_{t+1}$, and provides a reward $r_t$. The interaction with the environment stops after the $T$-th action $a_T$ is taken and the $T$-th reward $r_T$ is obtained. Transitions are typically assumed to be Markovian and  determined by the probability function $P^a_{s s^{'}} = \Pr{s_{t+1}=s'|s_t\!=\!s, a_t\!=\!a}$. There are two fundamental aspects of RL to be considered. The first one is that $r_t$ depends on $s_t$ and $a_t$ stochastically, whereas the second one is that the optimization (i.e., selection of the best action to take) is time-coupled. The reason for this being that action $a_t$ not only affects $r_t$, but also subsequent $s_{t'}$ for $t\!<t'\!\leq T\!$. This defines a time-coupled optimization problem that we frame using the formalism of the MDP.

In FH MDP problems, the goal is to learn a (non-stationary) policy $\pi$ that maximizes the expected cumulative reward $\mathbb{E}_{\pi}\big[\sum_{t=1}^Tr_t\big]$. Due to the lack of stationarity present in FH MDPs \cite{puterman2014markov}, this goal amounts to learning a set of policies $\pi=\{ \pi_t\}_{t=1} ^T$, where $\pi_t: \ccalS \mapsto \ccalA$ is a map from states to actions. The optimality of a policy is measured in terms of the expected rewards accumulated across $\ccalT=\{1,...,T\}$. Optimal policies define optimal actions, that lead to high cumulative rewards. In value-based methods, policies are learned indirectly from the VFs. In FH MDPs, the VF of a given state-action pair at time-step $t$ under a policy $\pi$ is the expected cumulative reward obtained by being in state $s_t$, taking the action $a_t$, and following a policy $\pi$ from time $t$ until the time horizon $T$. Formally, the VFs of a policy $\pi$ are defined as $Q^\pi =  \{ Q^\pi_t \}_{t=1}^T$, where $Q^\pi_t(s, a) = \mathbb{E}_\pi [ \sum_{t'=t}^T r_{t'} | s_t=s, a_t=a ]$. Relevantly, the optimality of the VFs holds element-wise, meaning that for every possible policy $\pi$, the optimal VFs fulfill the property $Q^*_t(s, a) \geq Q^\pi_t(s, a) \; \forall \; s, a, t$ \cite{puterman2014markov}\cite{sutton2018reinforcement}. The optimal VFs $Q^*=\{Q^*_t\}_{t=1}^T$ induce the optimal policy $\pi^*$, that can be obtained by maximizing greedily the VFs with respect to the actions $\pi^*_t(s) = \argmax_a Q^*_t(s, a)$.

When the model of the MDP (i.e., $P_{ss'}^a$ and $R_s^a$ for all $s,s',a$) is known, the optimal VFs can be obtained by recursively solving the non-linear \emph{Bellman optimality equations}:
\begin{subequations}\label{eq::bellman_equations_both}
\begin{align}
    \label{eq::bellman_equation}
    Q_t^*\left(s, a\right) &= R_s^a + \textstyle \sum_{s'} P_{ss'}^a \max_{a'} Q_{t+1}\left(s', a'\right), \\
    \label{eq::bellman_equation_terminal}
    Q_T^*\left(s, a\right) &= R_s^a.
\end{align}
\end{subequations}
\noindent The recursion, known as \emph{backward induction}, begins by obtaining first the VFs of the terminal stage $Q_T^*\left(s, a\right)$ for all $(s,a)$, which are trivially the values of the reward function [cf. \eqref{eq::bellman_equation_terminal}]. Then, the algorithm proceeds backward in time by maximizing over the VFs of the time-step previously computed.


RL deals with the pervasive case where the model of the MDP is either unknown or too cumbersome (large) to learn. The basic idea in RL is to estimate the optimal VFs directly from a set of \emph{transitions} \emph{sampled from the environment}. More specifically, one interacts $N$  times (with $N>>T$) with the environment, so that multiple FH trajectories of $T$ time-steps are run. In each of those interactions (indexed by $n=1,...,N$) the MDP is sampled to obtain the transition $\tau_n$, which comprises the tuple $\left( t_n, s_n, s_{n+1}, a_n, r_n \right)$. The elements $s_n$, $s_{n+1}$, and $r_n$ depend on the dynamics of the MDP ($P_{ss'}^a$ and $R_s^a$), while $a_n$ depends on a sampling policy $\tilde{\pi}$. Usually, policy $\tilde{\pi}$ guarantees high exploration right after initialization, and converges to the estimated (optimal) policy over time (e.g., epsilon-greedy policies \cite{bertsekas2019reinforcement}). Moreover, it is often the case that transitions are saved in a dataset $\ccalM_n=\{\tau_i\}_{i=1}^n$, so that they can be used later on to enhance the performance of the RL algorithm at hand.  $Q$-learning, the most popular value-based RL algorithm, has an FH variant \cite{vp2021finite} that proposes the following stochastic approximation to \eqref{eq::bellman_equations_both}:
\begin{subequations}\label{eq::q_learning}
\begin{align}    
    &Q_{t_n}^{n + 1}\!(s_n, a_n) \!=\! Q_{t_n}^{n}\!(s_n, a_n) \!+\! \alpha_n ( \hat{q}_n \!- \!Q_{t_n}^{n}(s_n, a_n)),~\text{with}\\
    &\hat{q}_n\!=\! r_n~\text{if}~t_n\!=\!T~\text{and}~\hat{q}_n\! = \!r_n \!+\! \max_a Q_{t_{n+1}}^n ( s_{n + 1}, a )~\text{if}~t_n\!<\!T;\nonumber\\
    &Q^{n + 1}_{t_n}(s, a)  = Q^n_{t_n}(s, a),~~\text{for~all}~(s,a)\neq(s_n,a_n);\\ 
    &Q^{n + 1}_{t}(s, a)  = Q^n_{t}(s, a),~~\text{for~all}~(s,a)~\text{and}~t\neq {t_n}; 
\end{align}
\end{subequations}
\noindent where $n$ is the iteration index, $\alpha_n$ a step-size, and $\hat{q}_t$ a target estimate.
Under technical conditions (including that $\alpha_n$ square summable but not summable, and that under $\tilde{\pi}$ all state-action pairs are sampled infinitely often), \textit{FHQ}-learning converges to the optimal VFs  \cite{bertsekas2012dynamic}\cite{vp2021finite}.

Unfortunately, when the state-action space is large, or when the time horizon is long, $Q$-learning suffers from the curse of dimensionality, since the number of entries of the VFs to estimate grows linearly with the number of states, the number of actions, and the time horizon. This is usually addressed by introducing a parametric approximator $Q_{t, \bbtheta_t}$ of the optimal VFs, with $\bbtheta_t$ being a vector of parameters that define the VF associated with the time instant $t$.  As the samples in $\ccalM_n$ are revealed sequentially, the estimation of $\{\bbtheta_t\}_{t=1}^T$ is usually formulated as an online optimization problem. In each iteration, the following quadratic problem is considered: 
\begin{align}
    \label{eq::q_approx}
    &\{\bbtheta_t^{n + 1}\}_{t=1}^T = \argmin_{\{\bbtheta_t\}_{t=1}^T} \sum_{i=1}^n \left( \hat{q}_i - Q_{{t_i}, \bbtheta_{t_i}} \left(s_i, a_i \right) \right)^2,~\text{with}\\
        &~\hat{q}_i\!= r_i~\text{if}~t_i=T~\text{and}~\hat{q}_i\! = \!r_i \!+\! \max_a Q_{t_{i+1},\bbtheta_{t_{i+1}}^n} \!( s_{i + 1}, a)~\text{if}~t_i\!<\!T.\nonumber
\end{align}
%
Since the complexity of the above problem grows with $n$, a (fully online) stochastic gradient descent approach is typically implemented, yielding to the following update rule:
\begin{subequations}\label{eq::q_approx_iteration}
\begin{align}
    \bbtheta^{n + 1}_{t_n}  &= \bbtheta^n_{t_n} 
    + \alpha_n \left( \hat{q}_n - Q_{{t_n}, \bbtheta^{n}_{t_n}} \left(s_n, a_n\right) \right) \nabla_{\bbtheta_{t_n}} Q_{{t_n}, \bbtheta^{n}_{t_n}} \left(s_n, a_n\right), \\
    \bbtheta^{n + 1}_{t'}  &= \bbtheta^n_{t'},~~\text{for}~t'\neq {t_n}. 
\end{align}
\end{subequations}
\noindent The $Q$-learning update defined in \eqref{eq::q_learning} is related to the update in \eqref{eq::q_approx_iteration}. In fact, \eqref{eq::q_learning} can be understood as a VF approximation problem where the parameters to estimate are directly the VFs. As in $Q$-learning, the stochastic updates in \eqref{eq::q_approx_iteration} are run leveraging samples obtained using a sampling policy $\tilde{\pi}$, which is highly exploratory at the beginning, but converges to the estimated policy over time.

Different models have been proposed in the literature. Linear and (non-linear) NN-based models have captured most of the attention. While the former exhibit some relevant theoretical guarantees, the latter lead to good empirical results. Alternatively, in this paper, we introduce a tensor low-rank algorithm to approximate the VFs. 

\section{Tensor low-rank approximation for FH VF}
\label{S:contribution}
\vspace{-2mm}
We propose a novel VF approximation technique for FH RL problems, that promotes low rank in a tensor representation of the state-action-time VFs. With this goal in mind, we first show how to model (represent) the VFs of an FH MDP as a tensor. We then formulate the tensor approximation problem and conclude by introducing a stochastic tensor low-rank algorithm. 

In discrete problems, the VFs are typically represented as matrices (aka tabular models). States are indexed in the rows and actions in the columns. Then, in FH problems, each time-step is associated with a VF matrix. However, as noted in \cite{rozada2023tensor} \cite{tsai2021tensor}, tensors (multi-way arrays) are a more natural representation of the VFs. State and action spaces are commonly multi-dimensional, so each dimension of the state-action space can be mapped to a dimension of a tensor of VFs. 
Hence, the first step in our approach is to model $Q$-functions as tensors. To that end, let $D_\ccalS$ and $D_\ccalA$ be the number of dimensions of the state-space $\ccalS$ and action-space $\ccalA$, respectively. Since states and actions are multidimensional, we represent them using the vectors $\bbs = \left[s_1, ..., s_{D_\ccalS} \right]^T$ and $\bba = \left[a_1, ..., a_{D_\ccalA} \right]^T$. Similarly, the state and action spaces can be written as $\ccalS = \ccalS_1 \times ... \times \ccalS_{D_\ccalS}$ and $\ccalA = \ccalA_1 \times ... \times \ccalA_{D_\ccalA}$, with $\times$ denoting the (set) Cartesian product and $\ccalS_i$ the domain of $s_i$, the $i$-th entry of the vector state. The cardinalities of the state and action spaces are $|\ccalS| = \prod_{i=1}^{D_\ccalS} |\ccalS_i|$ and $|\ccalA| = \prod_{j=1}^{D_\ccalA} |\ccalA_j|$. We define the joint state-action-time space $\ccalD = \ccalS \times \ccalA$, with $D = D_\ccalS + D_\ccalA$ dimensions, and cardinality $|\ccalD|=|\ccalS| |\ccalA|$. For the sake of notation clarity, we will denote the $d$-th dimension of $\ccalD$ as $\ccalD_d$, thus $|\ccalD|= \prod_{d=1}^{D} |\ccalD_d|$. The second step is to account for the FH horizon. As explained in the previous section, the lack of stationarity in FH RL environments implies that the VF is different for each $t=1,...,T$. Instead of handling this using $T$ different $Q$-tensors, we postulate one additional dimension for the $Q$-tensor, so that the number of dimensions is now $D+1$, with the last dimension indexing time (distance to the horizon). As a result, we represent VF associated with an FH RL setup by the tensor $\tenbQ \in \reals^{|\ccalD_1| \times ... \times |\ccalD_D|\times T}$, which contains $T|\ccalD|$ values. With this notation at hand, an entry of the $Q$-tensor $\tenbQ$ can be indexed as $\left[ \tenbQ \right]_{\left[\bbs ; \bba ; t \right]}$, where $t$ is the time index and $\bbs$ and $\bba$ are the vectors introduced at the beginning of this paragraph.

Although more natural, the tensor representation does not represent an advantage per se. We still have to estimate the $T|\ccalD|$ entries of $\tenbQ$, with the cardinality $|\ccalD|$ of the state-action-time space $\ccalD$ depending multiplicatively on the cardinality of each dimension $\ccalD_d$. To alleviate this problem we propose approximating $\tenbQ$ imposing a parsimonious (multilinear) tensor low-rank structure. As we show next, tensor low-rank approaches transform the multiplicative dependency into an additive one. 
More specifically, instead of using a parametric model $\{Q_{t, \bbtheta_t}\}_{t=1}^T$ to approximate the optimal VFs [see \eqref{eq::q_approx} in Sec. 2], we propose using a tensor low-rank \emph{non-parametric} approximation $\hat{\tenbQ}$. With this in mind, we propose solving the following online tensor completion problem [cf. \eqref{eq::q_approx}]:
\begin{align}
   &\hat{\tenbQ}^{n + 1} \!= \!\argmin_{\smalltenbQ} \sum_{i=1}^n ( \hat{q}_{t_i} \!-\! [ \tenbQ ]_{[\bbs_i ; \bba_i ; t_i ]} )^2~~\text{s.~to:}~\mathrm{rank}(\tenbQ)\!\leq K,\! \label{eq::tensor_lowrank}\\
           &~~~\text{with}~\hat{q}_i\! := \!r_i + \max_{\bba} [\hat{\tenbQ}^n]_{[\bbs_{i+1} ; \bba ; t_i+1 ]} ~\text{if}~t<T~~\text{and}~~\hat{q}_T\!:= r_T;\nonumber
\end{align}
\noindent where ``s.to'' stands for ``subject to'' and $K$ is the maximum rank of the tensor. The rank constraint in \eqref{eq::tensor_lowrank} is non-convex. To deal with this, we use the (truncated) PARAFAC decomposition \cite{bro1997parafac}. The $K$-rank PARAFAC decomposition of a $(D+1)$-dimensional tensor is defined as the sum of the outer product of vectors 
\begin{eqnarray}
    &\hat{\tenbQ} = \sum_{k=1}^K \bbq_1^k\circledcirc...\circledcirc\bbq_{D+1}^k.&
\end{eqnarray}
\noindent Interestingly, for all $d=1,...,{D+1}$, the  matrices $\hbQ_d=[\bbq_d^1,...,\bbq_d^K]$, known as factors, can be used to formulate the PARAFAC decomposition in matrix form. Specifically, we can matrizice the PARAFAC decomposition along each dimension, also called mode, as
\begin{equation}
    \label{eq::matrization}
    \mathrm{mat}_d(\hat{\tenbQ}) = (\hbQ_1 \odot...\odot \hbQ_{d-1} \odot \hbQ_{d+1} \odot ... \odot \hbQ_{D+1}) \hbQ_d^\top,
\end{equation}
\noindent where $\odot$ denotes the \emph{Khatri-Rao} product. Moreover, we define $\bigodot_{i \neq d}^{D+1} \hbQ_i:=\hbQ_1 \odot...\odot \hbQ_{d-1} \odot \hbQ_{d+1} \odot ... \odot \hbQ_{D+1}$ and the operator $\mathrm{unmat}_d(\cdot)$, the inverse of $\mathrm{mat}_d(\cdot)$, which given $\mathrm{mat}_d(\hat{\tenbQ})$ generates the output $\hat{\tenbQ}$.

Even after replacing the rank constraint in \eqref{eq::tensor_lowrank} with the (matricized) PARAFAC model in \eqref{eq::matrization}, we have to estimate the factors $\{ \hbQ_1, ..., \hbQ_{D+1}\}$, which is still a non-convex problem. Note, however, that optimizing over just one of the factors is convex. Thus, we propose using a block coordinate descent algorithm to approximate $\tenbQ^n$. Mathematically, at each interaction $n$, for each dimension $d$, we fix all other dimensions $j\neq d$ and consider the problem:
\begin{eqnarray}
    \nonumber
    \hbQ_d^{n + 1} &=& \argmin_{\bbQ_d} \sum_{i=1}^n \Big( \hat{q}_{t_i} - \Big[\hat{\tenbQ}_d \Big]_{[\bbs_i; \bba_i ; t_i]} \Big)^2\\
    &~&\text{s.~to}:~\hat{\tenbQ}_d = \mathrm{unmat}_d\Big(\Big(\bigodot_{j \neq d}^{D+1} \hbQ_j^n\Big) \bbQ_d^\top \Big). \label{eq::tensor_lowrank_parafac}
\end{eqnarray}
\noindent 

\vspace{-1mm}

Inspired by previous FH-LR algorithms, we reduce complexity by relying on stochastic gradients, leading to the proposed FH tensor low-rank algorithm (\textit{FHTLR-learning}), where, for each $n$, we run:
\begin{align}
    \label{eq::tensor_lowrank_parafac_iteration}
    &[\hbQ_d^{n + 1}]_{[\bbs_n;\bba_n; {t_n}]_d} = [\hbQ_d^n]_{[\bbs_n;\bba_n; {t_n}]_d}  \\ 
    &\hspace{0.3cm}+ \alpha_n \left( \hat{q}_{t_n} - [\hat{\tenbQ}_d^n]_{[\bbs_n;\bba_n; {t_n}]} \right)\! \big[\hbQ_{\backslash d}^n\big]_{[\bbs_n;\bba_n;{t_n}]}, ~\text{for}~d\!=\!1,...,D\!+\!1\nonumber
\end{align}
\noindent where each term in the above expression is defined as follows:
\begin{itemize}[leftmargin=*]
\item $[\hbQ_d^n]_{[\bbs_n;\bba_n; {t_n}]_d} \in \reals^K$ is a vector whose entries are the row of matrix $\hbQ_d^n$ indexed by the $d$-th entry of the  vector $[\bbs_n;\bba_n; {t_n}]$.
\item $\hat{q}_{t_n}$ is the current target estimate of the VF, which is set to 
\begin{equation}\label{eq::tensor_lowrank_parafac_iteration_term1}
\hat{q}_{t_n}\! = \!r_n + \max_{\bba} [\hat{\tenbQ}^n]_{[\bbs_{n+1} ; \bba ; t_n+1 ]}~\text{if}~t_n<T~~\text{and}~~\hat{q}_T= r_T.
\end{equation}
\item $\hat{\tenbQ}_d^n$ is the tensor obtained unmatricizing along the $d$-th dimension:
\vspace{-1mm}
\begin{equation}\label{eq::tensor_lowrank_parafac_iteration_term2}
\hat{\tenbQ}_d = \mathrm{unmat}_d\Big(\Big(\bigodot_{j \neq d}^{D+1} \hbQ_j^n\Big) \bbQ_d^\top \Big).
\end{equation}
\item $\big[\hbQ_{\backslash d}^n\big]_{[\bbs_n;\bba_n; t_n]} \in \reals^K$ is a  vector whose $k$-th entry is 
\begin{equation}\label{eq::tensor_lowrank_parafac_iteration_term3}
\Big[\big[\hbQ_{\setminus d}^n\big]_{[\bbs_n;\bba_n; t_n]}\Big]_k = \!\prod_{j\neq d}^{D+1} \!\Big[[\hbQ_{j}^n]_{[\bbs_n;\bba_n; t_n]_j}\Big]_k.
\end{equation}
\end{itemize}
\noindent

\vspace{-1mm}

\textit{FHTLR-learning} works as follows. In each time-step $n$ we sample a transition $\tau_n=\left(t_n, s_n, s_{n+1}, a_n, r_n \right)$ using a sampling policy $\tilde{\pi}$. For each dimension $d=1,...,{D+1}$ of the estimated tensor $\hat{\tenbQ}$, we fix the remaining dimensions, then we update the corresponding entries using the update rule in \eqref{eq::tensor_lowrank_parafac_iteration}, where each of the terms in the expressions can be obtained using the previous estimate along with \eqref{eq::tensor_lowrank_parafac_iteration_term1}-\eqref{eq::tensor_lowrank_parafac_iteration_term3}. We repeat for all dimensions and increase the index $n$ by one. In our particular FH context, the $t$-th row of matrix $\hbQ_{D+1}$ represents the embedding of the $t$ time instant into the $K$-dimensional domain of the low-rank $Q$-tensor representation. As a result, if two time instants $t$ and $t'$ have similar embbedings, their associated VFs $\hat{Q}_t(\bbs,\bba)$ and $\hat{Q}_{t'}(\bbs,\bba)$ are expected to be close for all $(\bbs,\bba)$ as well. 

\begin{figure*}[!t]
    \begin{minipage}{0.48\textwidth}
        \centering
        \begin{minipage}{\linewidth}
        \centering
        \begin{tabular}{l|l|r|r}
            \hline
            \textbf{Environments} & \textbf{Algorithms} & \textbf{Return} & \textbf{$\#$ Params.} \\
            \hline
            \hline
            \multirow{3}{*}{Gridworld} & \textit{Q}-learning & 78.35 & 100 \\
            \cline{2-4}
            & \textit{FHQ}-learning & 89.10 & 500 \\
            \cline{2-4}
            & \textit{FHTLR}-learning & 89.10 & 152 \\
            \hline
            \hline
            \multirow{4}{*}{Wireless} & \textit{FHQ}-learning & -1.23 & 20,000,000 \\
            \cline{2-4}
            & \textit{DQN} & 0.96 & 3,524 \\
            \cline{2-4}
            & \textit{DFHQN} & 1.38 & 16,724 \\
            \cline{2-4}
            & \textit{TLR}-learning & 1.36 & 3,450 \\
            \hline
        \end{tabular}
        \captionof{table}{Mean returns and number ($\#$) of parameters.}
        \label{tab::results}
        \end{minipage}

        \begin{minipage}{\linewidth}
        \centering
        \includegraphics[width=\linewidth]{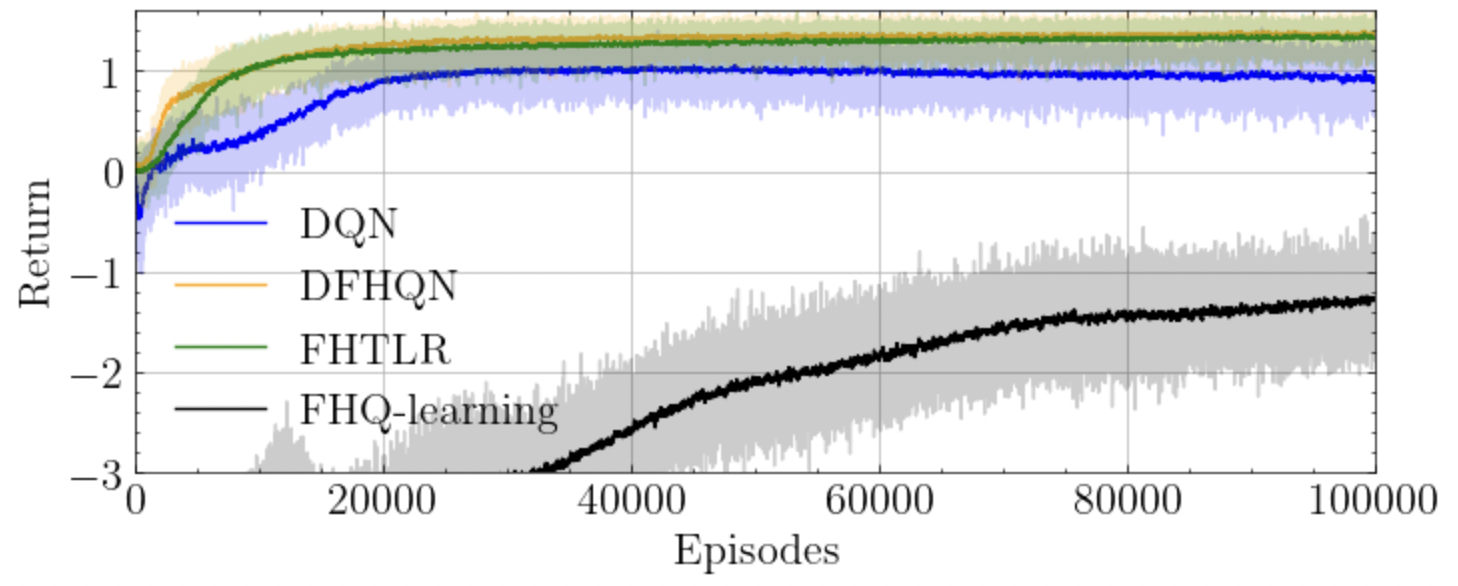}
        \caption{Mean return per episode in the wireless-communications environment. \textit{FHTLR}-learning and \textit{DFHQN} achieve a similar return, and outperform \textit{DQN}.}
        \label{fig:exp_communications}
        \end{minipage}
    \end{minipage}%
    \hspace{0.03\textwidth}
    \begin{minipage}{0.48\textwidth}
        \centering
        \includegraphics[width=0.84\linewidth]{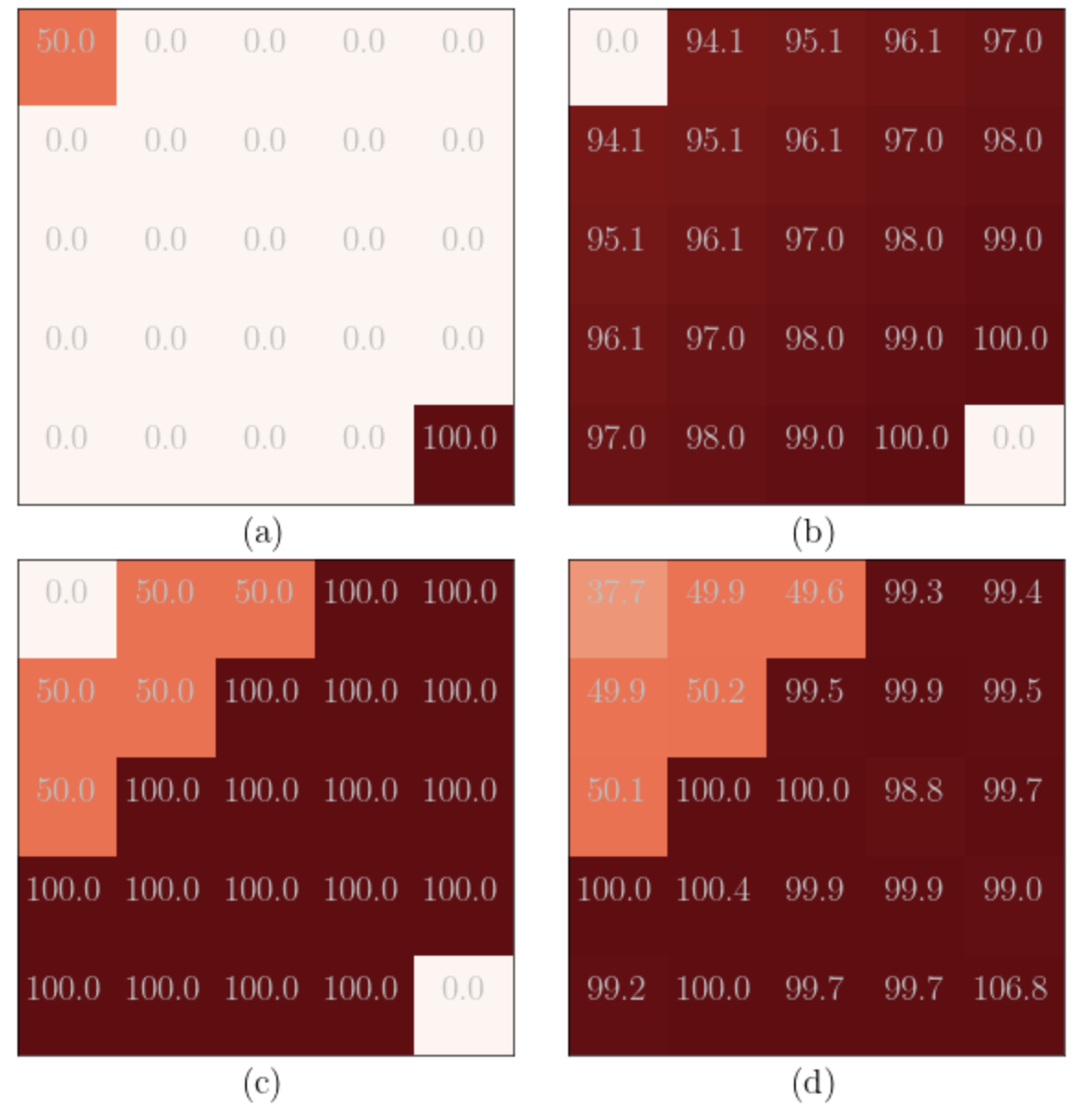}
        \caption{Picture (a) shows the rewards of the grid-world environment. The remaining pictures show in time-step $0$ the estimated VFs by (b) $Q$-learning, (c) \textit{FHQ}-learning, and (d)   \textit{FHTLR}-learning.}
       \label{fig:exp_gridworld}
    \end{minipage}
\end{figure*}

\textit{FHTLR-learning} reduces significantly the number of parameters that we need to estimate. The $K$-rank PARAFAC model has ${D+1}$ factors (one per dimension), being each factor $\hbQ_d$ of size $|\ccalD_d| \times K$. Thus, the total number of parameters of the PARAFAC model is $ (T+\sum_{d=1}^D\left|\mathcal{D}_d\right|) K $. This contrasts with the number of entries of the original tensor, which is $T|\ccalD|= T\prod_{d=1}^{D+1} |\ccalD_d|$. Then, with our approach, enlarging the time horizon $T$ has an additive effect on the number of parameters to estimate, instead of a multiplicative one. 

\vspace{-4mm}

\section{Numerical experiments}
\label{S:simulations}

\vspace{-3mm}

We tested empirically \textit{FHTLR}-learning against $Q$-learning and deep $Q$-learning (\textit{DQN}), and their FH versions \textit{FHQ}-learning and \textit{DFHQN} \cite{de2020fixed}, which is the state-of-the-art method in FH value-based problems. We simulated two scenarios: a simple (more didactic) grid-like problem, and a more challenging wireless communications problem. The implementation details can be found in the code repository \cite{rozada2023}.

\vspace{1mm}

\noindent \textbf{Grid world environment.} This setup is a simple $5 \times 5$ grid-like problem used to illustrate how \textit{FHTLR}-learning helps overcoming the challenges of FH problems. The time horizon is $T=5$ time-steps, and there are two rewards of $50$ and $100$ placed in opposite corners of the grid (see Fig. \ref{fig:exp_gridworld}). The initial state is selected uniformly at random. In an infinite-horizon scenario, the optimal policy consists of moving towards the high-reward goal. However, in FH problems, it is not always feasible to reach the high-reward goal in the remaining time-steps. Then, upon initialization, the optimal policy would occasionally be moving towards the low-reward goal. We have compared regular $Q$-learning against   \textit{FHQ}-learning, and \textit{FHTLR}-learning. The results are depicted in Fig. \ref{fig:exp_gridworld} and summarized in Table \ref{tab::results}. Notably, traditional (non-FH) $Q$-learning fails to approximate the optimal VFs, resulting in suboptimal return outcomes. Both \textit{FHTLR}-learning and \textit{FHQ}-learning yield high return results. However, \textit{FHTLR}-learning estimates properly the optimal VFs using  significantly less amount of parameters compared to \textit{FHQ}-learning. 

\vspace{1mm}

\noindent \textbf{Wireless communications environment.} This setup presents a time-limited opportunistic multiple-access wireless setup. A single user, equipped with a battery, and a queue, transmits packets to the access point. There are $C=2$ orthogonal channels and transmissions must occur during a finite period of $T=5$ time-steps. The user access opportunistically and the channel may be occupied or not. The state of the system is given by: a) the fading level and the occupancy state of each of the $C=2$ channels, b) the energy in the battery, and c) the number of packets in the queue.  In each time-step, the user observes the state and selects the (discretized) power to send through each of the channels. The rate of transmission is given by Shannon's capacity formula. If the channel is occupied, there is a packet loss of $50\%$. The reward is $0$ in all time-steps but in the terminal one, where the reward is a weighted sum of the battery level (positive weight) and the remaining number of packets in the queue (negative weight). Thus, the agent needs to trade off throughput and battery, with time playing a central role. In the initial time-steps, the agent can be conservative and wait for high-SNR-free channels to transmit. However, as time passes, clearing the queue becomes more relevant. This leads to less efficient transmissions that consume the battery. Additional details on the system parameters as well as the random dynamics for the occupancy, fading, energy harvesting, and packet arrivals are detailed in \cite{rozada2023}. We compared the average performance of \textit{FHQ}-learning, \textit{DQN}, \textit{DFHQN}, and \textit{FHTLR}-learning in this setup over $100$ experiments. The results are shown in Fig. \ref{fig:exp_communications} and in Table \ref{tab::results}. Due to the number of $Q$-values to estimate, \textit{FHQ}-learning needs far more samples to achieve good results. \textit{DFHQN}, and \textit{FHTLR}-learning consistently achieve higher rewards than \textit{DQN}, which estimates a (stationary) policy unable to perform at the same level. However, \textit{DFHQN} needs roughly $T=5$ times more parameters than \textit{DQN}. Notably, \textit{FHTLR}-learning achieves similar returns with approximately $20 \%$ of the size of \textit{DFHQN}.

\vspace{-3mm}

\section{Conclusions}
\label{S:conclusions}

\vspace{-2mm}

This paper presented an FH tensor low-rank algorithm (\textit{FHTLR}-learning) to estimate the optimal VFs of an FH MDP. Specifically, we modeled the VFs as a multi-dimensional tensor, with time being one of the dimensions of the tensor. We then used the PARAFAC decomposition to leverage low-rank in a tensor completion problem to approximate the VFs. As the size of the PARAFAC model grows additively in the dimensions of the tensor, our approach is well-suited to keep the number of parameters to estimate under control. We tested the proposed design in two RL environments (a grid-world-like setup, and a wireless communications setup) and compared it with \textit{FHQ}-learning, and \textit{DFHQN}. \textit{FHTLR}-learning leads to high returns while being much more efficient in terms of use of parameters. 

\newpage
\bibliographystyle{IEEEtran}
\bibliography{references}

\end{document}